\pdfoutput=1
\documentclass{article}

\usepackage{preprint}

\usepackage[utf8]{inputenc}
\usepackage[T1]{fontenc}
\usepackage{microtype}
\usepackage[hyphens]{url}
\usepackage{graphicx}
\usepackage{booktabs}
\usepackage{amsmath,amssymb,amsfonts}
\usepackage{nicefrac}
\usepackage[numbers,sort&compress]{natbib}
\usepackage{caption}
\usepackage{subcaption}
\usepackage[dvipsnames]{xcolor}
\usepackage{tikz}
\usetikzlibrary{positioning, arrows.meta, fit, shapes.geometric, calc, shadows}
\usepackage{algorithm}
\usepackage{algorithmic}

\definecolor{linkblue}{RGB}{0,82,155}
\usepackage[colorlinks=true,linkcolor=linkblue,citecolor=linkblue,urlcolor=linkblue]{hyperref}

\title{Chain Dimension Scaling: Communication-Efficient\\ Loop-Style Scaling for Industrial Short-Video Ranking}

\author{%
  Zhimeng Jiang \quad Yuening Li \quad Ruixiao Sun \quad Yuanzhen Lin \\
  Kun Jin \quad Chuan He \quad Liang Liu \\[0.6ex]
  \affil{Google}%
}

\begin{document}

\maketitle

\begin{abstract}
Industrial recommendation rankers have continued to scale in width and depth with billions of parameters to improve modeling capacity, but this trend increasingly incurs substantial costs in parameter growth, distributed communication overhead, and training inefficiency. In this work, we introduce \underline{C}hain \underline{D}imension \underline{S}caling (CDS), a scaling strategy that is orthogonal to conventional width and depth scaling. CDS repeatedly applies a single feature-interaction block during the forward pass while sharing parameters across iterations, where the number of iterations (i.e., the chain length) serves as a novel scaling dimension without introducing extra model parameters. We further incorporate a lightweight intra-chain residual connection to stabilize optimization and improve model quality.
CDS naturally enables two operating regimes. In the \emph{compression regime}, shallow blocks combined with a longer chain can match the quality of deeper baseline models while reducing parameter count and improving training efficiency; in the \emph{quality-scaling} regime, increasing the chain length under a fixed parameter budget leads to consistent gains in both offline and online ranking performance, suggesting that the chain dimension provides additional modeling capacity beyond parameter reuse alone. We evaluate CDS in a large-scale multi-task short-video ranking system in production. In the compression regime, CDS maintains the quality of offline ranking (within $0.03\%$ of the baseline) while reducing the trainable parameters by 26.6\% and improving training throughput by 6.4\%, with neutral-to-positive online participation. In the quality-scaling regime, we observe monotonic quality gains as chain length increases under a fixed
parameter budget. Further wall-clock analysis reveals that the efficiency improvement mainly comes not from reduced FLOPs, but from lower cross-device synchronization overhead during distributed training and reduced optimizer update cost. These results position parameter-shared iterative computation as a practical and communication-efficient scaling axis for industrial recommendation ranking systems.
\end{abstract}

\section{Introduction}\label{sec:intro}
Industrial recommendation rankers have grown rapidly in recent years, scaling to billions of parameters to improve ranking quality. Existing scaling strategies mainly rely on larger hidden dimensions, deeper feature-interaction stacks, and larger-scale training data~\citep{wang2021dcnv2,song2019autoint,zhai2024hstu,han2025mtgr,zhang2025onetrans,chen2019bst,zhu2025rankmixer,khrylchenko2025billion,zhang2024wukong}. Similar to language modeling~\citep{kaplan2020scaling,hoffmann2022chinchilla}, these scaling axes consistently improve model quality, but also introduce substantial training and serving costs.

The cost of scaling is not limited to FLOPs or parameter count. In modern distributed training systems using sharded data parallelism such as ZeRO-3~\citep{rajbhandari2020zero} and
FSDP~\citep{zhao2023fsdp}, \textbf{every additional independently parameterized layer triggers an all-gather of its sharded parameters
in the forward pass and a reduce-scatter of its gradients in the
backward pass}. As ranking models become deeper, these synchronization costs grow significantly and increasingly dominate wall-clock training time in production systems. This makes communication overhead an important bottleneck for large-scale recommendation ranking.

In our production short-video ranker, the stacked feature-interaction module dominates both parameter count and training cost. Importantly, these costs scale with the number of independently parameterized layers rather than the effective depth of computation. This observation motivates a simple question: \emph{can we increase effective computation depth without increasing the number of unique parameterized layers?}

To answer this question, we propose \textbf{Chain Scale Dimension (CDS)}, a scaling strategy based on iterative computation with shared parameters. CDS repeatedly applies a single feature-interaction block during the forward pass while sharing parameters across iterations. The number of iterations, referred to as the \emph{chain length}, introduces a new scaling dimension that is orthogonal to width, depth, and data scaling. Since parameters are shared across iterations, increasing chain length does not increase trainable parameter count. 

Weight reuse has been explored in several areas, including recurrent Transformers~\citep{dehghani2018universal}, ALBERT~\citep{lan2020albert}, and recent loop-style language models~\citep{saunshi2025looped,geiping2025recurrent,zhu2025loopedlm,zhu2025vld}. Concurrent work has also begun exploring similar ideas for CTR prediction~\citep{tang2026loopctr}. Our work focuses on a different aspect of the problem: at industrial scale, the main benefit of parameter sharing is not reduced FLOPs, but reduced communication overhead under sharded distributed training~\citep{rajbhandari2020zero,zhao2023fsdp}. This systems perspective becomes increasingly important as recommendation models continue to scale.

CDS exposes two distinct operating regimes. In the \textbf{compression regime}: a smaller backbone combined with a longer chain can match baseline quality with fewer parameters and faster training. In the \textbf{quality-scaling regime}, increasing chain length under a fixed parameter budget consistently improves ranking quality, showing that the chain dimension provides additional modeling capacity beyond simple parameter reuse. 

Beyond these two regimes, a key efficiency finding emerges almost as a byproduct, and turns out to be the most generalizable. We monitor the training step end-to-end and find that the speedup of the configuration relative to the production baseline does \textbf{not} come from FLOPs reduction \footnote{forward FLOPs are essentially unchanged} but from two sources that prior loop-style work has not surfaced: fewer parameter
all-gather and gradient reduce-scatter operations across the smaller
number of \emph{unique} parameterized layers, and a smaller optimizer
step because there are fewer parameters to update. Parameter-shared iterative computation is therefore best understood not as a FLOPs-efficiency technique but as a \textbf{communication-efficiency} technique under modern sharded data-parallel training. We believe this framing applies wherever modern data-parallel rankers scale to many parameterized layers, well beyond the specific instantiation here.

To stabilize optimization under long chains, we introduce a lightweight \textbf{intra-iteration residual} design inspired by prior work on weight-tied deep networks~\citep{zhang2019fixup,bachlechner2021rezero}. We evaluate CDS in a large-scale multi-task short-video ranking system. In the compression regime, CDS reduces trainable parameters by $26.6\%$ and improves training throughput by $6.4\%$ while maintaining offline ranking quality with less than a $0.03\%$ degradation, together with neutral-to-positive online engagement improvements. In the quality-scaling regime, ranking quality improves monotonically as chain length increases under a fixed parameter budget.

Our contributions are summarized as follows:
\begin{itemize}
    \item We propose \textbf{Chain Dimension Scaling (CDS)}, a parameter-shared iterative scaling strategy that introduces chain length as a scaling dimension orthogonal to width, depth, and data.
    \item We identify two operating regimes of CDS: a \textbf{compression regime} that improves training efficiency at matched quality, and a \textbf{quality-scaling regime} where ranking quality improves monotonically with chain length under a fixed parameter budget with intra-iteration residual.
    \item We evaluate CDS in a production multi-task short-video ranker, where it reduces trainable parameters by $26.6\%$, improves training throughput by $6.4\%$, and achieves neutral-to-positive online engagement impact.
    \item We provide a wall-clock analysis showing that the efficiency gain mainly comes from reduced cross-device communication and optimizer overhead, highlighting parameter-shared iterative computation as a \textbf{communication-efficient} scaling strategy for industrial recommendation systems.
\end{itemize}

\section{Preliminaries}
\label{sec:prelim}
 
\subsection{Notations}
\label{sec:prelim:ranker}
We consider a multi-task recommendation ranker. The input feature layer
produces a token sequence $\mathcal{T} = (t_1, \dots, t_N) \in \mathbb{R}^{N \times d}$, where
each token $t_n$ represents the embedding of an input feature group (dense
features, sparse features, short- and long-term sequential user
history), all feature groups are projected into a shared embedding dimension $d$. The ranker contains a stack of $L$ feature-interaction blocks with independent parameters $W^{(\ell)}$:
\begin{equation}
  h^{(\ell)} = \text{Block}\!\left(h^{(\ell-1)};\, W^{(\ell)}\right),
  \qquad \ell = 1, \dots, L,
  \label{eq:baseline-stack}
\end{equation}
with $h^{(0)} = \mathcal{T}$. The output $h^{(L)}$ feeds a
shared projection layer followed by $K$ task-specific towers that produce
predictions $\hat{y}_1, \dots, \hat{y}_K$ for the multi-task objectives
the ranker optimizes (e.g., engagement, satisfaction). For example, each block can be implemented using a pointwise attention module from the HSTU~\citep{zhai2024hstu}; The discussion in this paper, however, is independent of the specific block design and only assumes a generic feature-interaction module.
 
\subsection{Where the Cost Concentrates}
\label{sec:prelim:bottleneck}
 
\begin{figure}[t]
  \centering
  \includegraphics[width=0.85\linewidth]{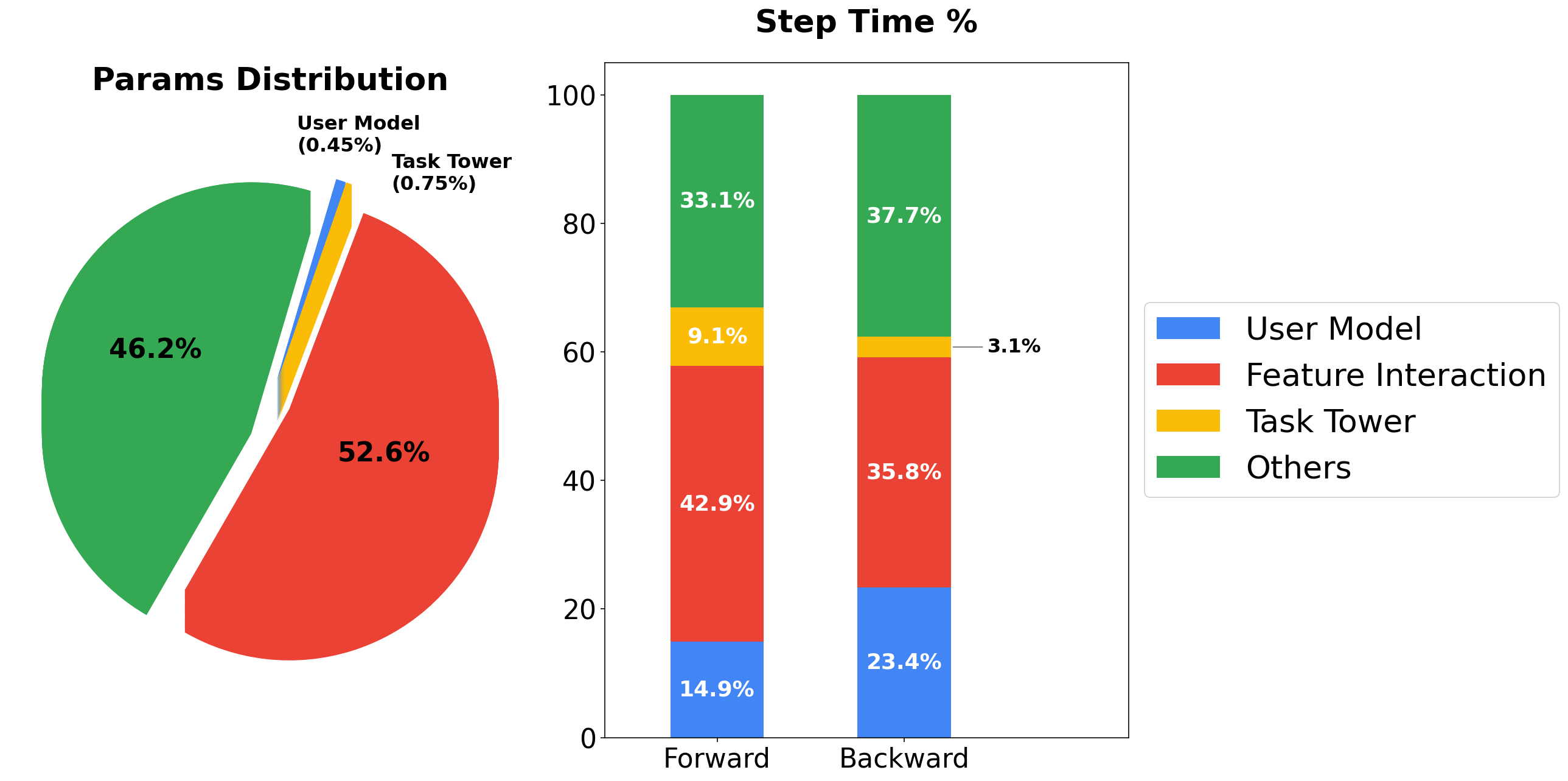}
  \caption{The ratio of trainable parameters (left) and training-step
  wall-clock time (right) across the major components of the
  production ranker. The stacked feature-interaction module dominates
  both, motivating it as the target for the new scaling axis.}
  \label{fig:bottleneck}
\end{figure}

Figure~\ref{fig:bottleneck} shows the parameter distribution and training-step wall-clock breakdown
of the ranker model. The stacked
feature-interaction module accounts for the majority of both trainable parameters and training step time, while the remaining
components---the input feature layer, user modeling modules, and the
multi-task towers---contribute relatively little by comparison.
 
Two observations follow from this result. First, any meaningful improvement in model efficiency
primarily come from optimizing the feature-interaction bottleneck module; Second,
the module is also the main communication bottleneck under sharded data-parallel training~\citep{rajbhandari2020zero,zhao2023fsdp}. Since the model consists of $L$ independently parameterized blocks, each layer introduces parameter all-gather operations in the forward pass and gradient all-reduce operations in the backward pass. As the model depth increases, these communication costs scale accordingly. This concentration of both parameter and communication cost motivates the scaling strategy introduced in the next section.

\section{Method}
\label{sec:method}
This section presents Chain Dimension Scaling (CDS). We first motivate the
new axis from the cost structure discussed in
Section~\ref{sec:prelim:bottleneck}
(Section~\ref{sec:method:positioning}). We then define the CDS forward pass
(Section~\ref{sec:method:cds}), introduce the intra-iteration residual
that design for better model quality (Section~\ref{sec:method:residual}), and finally discuss the two operating regimes enabled by the chain dimension (Section~\ref{sec:method:regimes}).
 
\subsection{From Cost Structure to a New Axis}
\label{sec:method:positioning}
 Section~\ref{sec:prelim:bottleneck} showed that the stacked feature-interaction module is the main bottleneck for both parameter footprint and distributed training cost. Stacking $L$ independently parameterized blocks increases parameter count, optimizer state size, and cross-device communication overhead under sharded data-parallel training~\citep{rajbhandari2020zero,zhao2023fsdp}. In particular, each additional layer introduces parameter synchronization (all-gather operations) in the forward pass and gradient all-reduce operations in the backward pass. These costs scale with the number of independently parameterized layers rather than the effective depth of computation.
 
Existing scaling strategies for industrial recommendation rankers primarily scale along width, depth, and data axes~\citep{zhai2024hstu,han2025mtgr,zhang2025onetrans,zhu2025rankmixer,khrylchenko2025billion,zhang2024wukong},
While effective, these approaches increase parameter count and distributed communication cost together. This motivates a simple question: \emph{can effective computation depth be increased without increasing the number of independently parameterized layers? }If so, the resulting scaling strategy could improve model capacity inside the feature-interaction module while avoiding much of the additional parameter, optimizer, and communication overhead.
 
\subsection{Chain Dimension Scaling}
\label{sec:method:cds}
We introduce a new scaling axis inside the feature-interaction module. Instead of stacking $L$ independently
parameterized blocks in Eq.~\ref{eq:baseline-stack}, CDS uses a shallow stack
of $L'$ ($L'<L$) \emph{shared-parameter} blocks, where each block is applied iteratively for $C$ steps. 

Let $\mathcal{B}_{\theta_\ell}$ denote a feature-interaction block with
parameters $\theta_\ell$. The CDS forward pass is defined as:
\begin{equation}
  h^{(\ell, c)}
  \;=\;
  \mathcal{B}_{\theta_{\ell}}\!\left(h^{(\ell, c-1)}\right),
  \qquad
  c = 1, \dots, C,
  \quad
  \ell = 1, \dots, L',
  \label{eq:cds-forward}
\end{equation}
where $h^{(\ell, 0)} = h^{(\ell - 1, C)}$ is the previous block's
final-iteration output, $h^{(1, 0)} = \mathcal{T}$, and the final interaction
module output $h^{(L', C)}$ feeds the downstream projection layer and task-specific
towers. We refer to $C$ as the \textbf{chain length} and the
corresponding scaling axis as the \textbf{chain dimension}.

\begin{figure*}[t]
  \centering
  \includegraphics[width=0.95\textwidth]{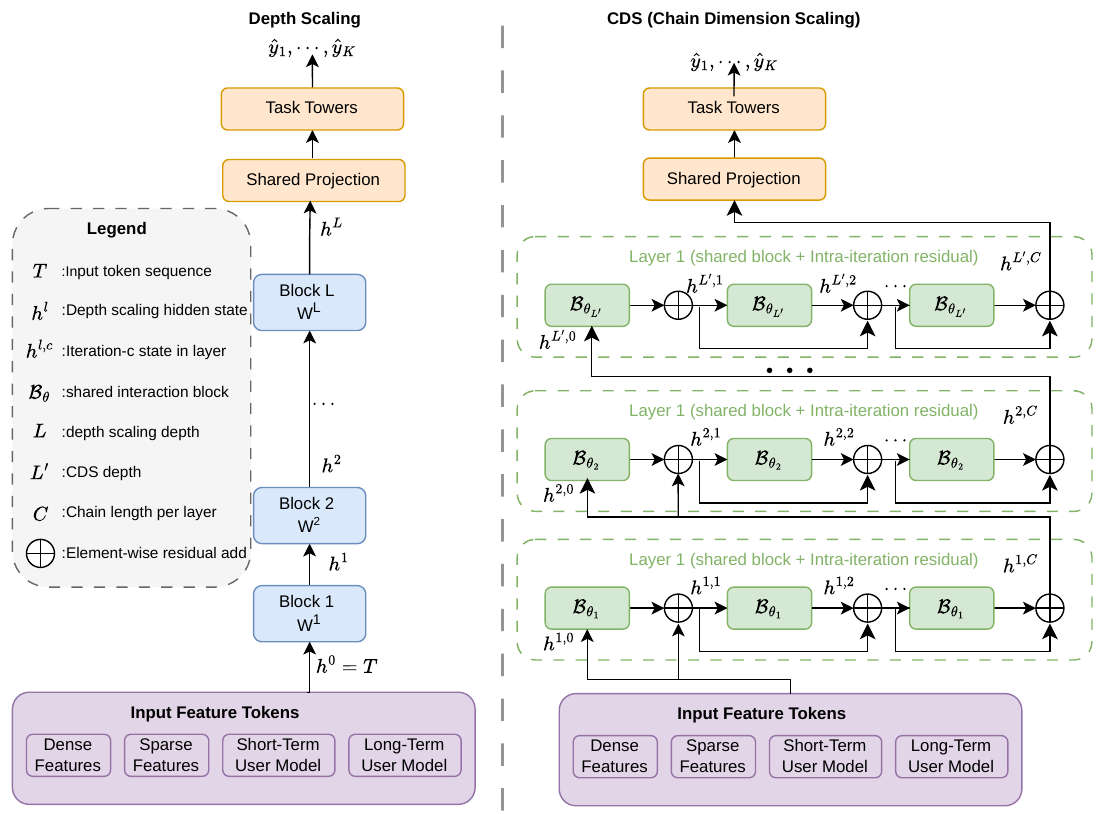}
  \caption{Architecture comparison betweendepth scaling and CDS.}
  \label{fig:architecture}
\end{figure*}

Figure~\ref{fig:architecture} compares the baseline architecture with CDS. In the baseline, the interaction module consists of $L$ independent blocks. In CDS, the module contains only $L'$ unique blocks, each reused across $C$ iterations.

The chain dimension has several useful properties. (1) It increases effective computation depth without increasing parameter count. The effective depth becomes $L' \cdot C$, while the number of trainable parameters depends only on $L'$.
(2) Under sharded data-parallel training~\citep{rajbhandari2020zero,zhao2023fsdp}, communication overhead scales with the number of unique parameterized layers. Increasing $C$ at fixed $L'$ therefore adds computation depth without introducing additional parameter synchronization or gradient all-reduce operations. (3) The chain dimension is orthogonal to conventional scaling axes such as width, depth, and data size. Setting $C=1$ recovers the standard stacked architecture, while larger values of $C$ move the model along the new scaling axis.
 
Because all iterations preserve tensor shape, CDS can be integrated into existing rankers with minimal architectural changes. The input feature layer, shared projection layer, and task-specific towers remain unchanged. Parameter sharing has previously been explored in architectures such as ALBERT~\citep{lan2020albert}, where parameters are tied across layers. CDS differs in that it introduces chain length as an explicit scaling knob that can be adjusted independently of parameter count. As we show later, different operating points along this axis lead to different trade-offs between model quality and system efficiency.
 
\subsection{Stabilizing Long Chains: The Intra-Iteration Residual}
\label{sec:method:residual}
Naive iterative parameter sharing is often difficult to optimize, especially at larger chain lengths~\citep{zhang2019fixup,bachlechner2021rezero}. We observe the same behavior in our setting: directly applying Eq.~\ref{eq:cds-forward} with $C \geq 2$ underperforms the corresponding baseline at matched effective depth.

To stabilize optimization, we introduce a lightweight \textbf{intra-iteration residual} connection:
\begin{equation}
  h^{(\ell, c)}
  \;=\;
  \mathcal{B}_{\theta_{\ell}}\!\left(h^{(\ell, c-1)}\right)
  \;+\;
  h^{(\ell, c-1)},
  \qquad
  c = 1, \dots, C,
  \label{eq:intra-skip}
\end{equation}
where $\mathcal{B}_{\theta_{\ell}}$ denotes
feature-interaction block with parameterized by $\theta_l$.
The residual introduces an exact identity path with each iteration, improving gradient propagation and stabilizing training at longer chain lengths.

We compare several residual designs in
Section~\ref{sec:experiments}. The intra-chain residual adds a skip connection inside each iteration. We also consider an
\emph{inter-iteration} residual connects only the chain's input and output of the entire chain:
\begin{equation}
  h^{(\ell, c)}
  \;=\;
  \mathcal{B}_{\theta_{\ell}}\!\left(h^{(\ell, c-1)}\right)
  \;+\;
  h^{(\ell, 0)}\qquad
  c = 1, \dots, C.
  \label{eq:inter-skip}
\end{equation}
 A \emph{combined} variant intra-inter residual is also investigated. Empirically,
the intra-iteration residual provides the largest improvement in optimization stability and ranking quality. The
inter-iteration residual alone has limited quality benefit and introduces additional training overhead, while combining both residual designs does not further improve performance over the intra-chain residual alone.
 
\subsection{Two Operating Regimes Along the Chain Dimension}
\label{sec:method:regimes}
 
Equation~\ref{eq:cds-forward} introduces a new scaling axis through the chain length $C$. In practice, this axis supports two different operating regimes, which we study throughout the experiments.

\paragraph{Compression regime.}
We first consider the setting where the goal is to match the quality of a baseline model with lower training cost. Starting from a baseline with depth $L$ and $C=1$, we
choose a shallow parameterized depth $L' < L$ and increase the chain length $C > 1$ such that the effective depth $L' \cdot C=L$
is preserved. Since the parameter count is determined by $L'$ alone,
this configuration reduces the number of trainable parameters while maintaining
similar effective computation depth. If model quality is preserved, the result is a more parameter- and communication-efficient architecture.
 
\paragraph{Quality-scaling regime.}
We next study whether the chain dimension can improve model quality under a fixed parameter budget. n this setting, we fix the parameterized depth $L'$ and increase the chain length $C$. Each increase in $C$ adds effective computation depth without introducing additional trainable parameters. If the chain dimension provides meaningful modeling capacity, ranking quality should improve as $C$ increases.
 
The two regimes serve different purposes. The compression regime demonstrates the practical system benefit of CDS, while the quality-scaling regime evaluates whether the chain dimension behaves as a genuine scaling axis rather than simply acting as a form of parameter sharing or regularization. We analyze the corresponding wall-clock training behavior in Section~\ref{sec:experiments:rq4}.

\section{Experiments}\label{sec:experiments}
In this section, we conduct extensive experiments to answer the following research questions:
\textbf{(RQ1) Compression:} Can CDS reduce parameters and training cost
while maintaining ranking quality? \textbf{(RQ2) Scaling:} Does increasing chain length improve ranking quality under a fixed parameter budget?
\textbf{(RQ3) Residual Design:} Which residual design best stabilizes training for long chains?
\textbf{(RQ4) Efficiency Mechanism:} Where does the training speedup of CDS come from?

\subsection{Experiment Setup and Evaluation Metrics}
\label{sec:experiments:setup}
 
We evaluate CDS on a large-scale short-form video (SFV) recommendation platform serving billions of ranking requests per day. The ranker is trained on user
interaction logs from the our system and jointly optimizes multi-task objectives spanning engagement and
satisfaction; Several tail tasks have positive label rates on the order of
$10^{-4}$ to $10^{-5}$, which are handled through the standard multi-task
loss weighting used by the production system. Models are implemented
in our internal recommendation framework and trained with sharded data-parallel
training~\citep{rajbhandari2020zero,zhao2023fsdp,xu2021gspmd} on a distributed
accelerator cluster.
 
We conduct extensive offline and online A/B experiments for every
configuration. For offline evaluation, we measure model quality by
AUC for classification tasks. For online evaluation, we run 10-day A/B tests in which a portion of real production traffic is allocated to both control and treatment models.

\subsection{RQ1: Compression Regime and Online A/B Validation}
\label{sec:experiments:rq1}
\begin{figure}[t]
  \centering
  \includegraphics[width=0.7\linewidth]{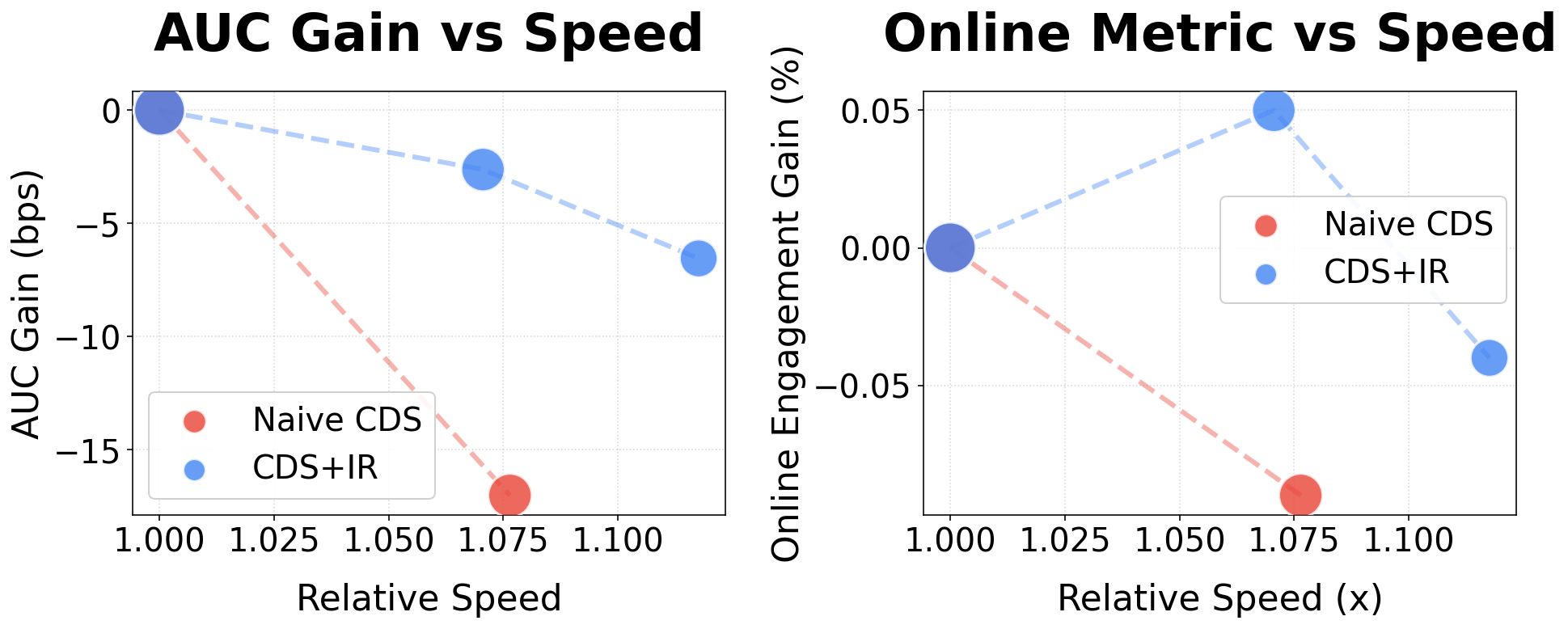}
  \caption{Pareto comparison of compression strategies.
  \emph{Naive} reduces the number of independently parameterized
  blocks without a residual connection;
  \emph{CDS+IR} is CDS with the intra-iteration residual.
  Dot size is proportional to trainable parameter count.
  Left: offline AUC gain over the production baseline (bps).
  Right: online engagement delta. CDS dominates the naive variant at
  every parameter budget on both metrics.}
  \label{fig:rq1_pareto}
  \vspace{-2mm}
\end{figure}

We first evaluate CDS in the compression regime. The goal is to reduce parameter count and training cost while maintaining model quality. We compare CDS against both the production baseline and a naive compression strategy that simply removes independently parameterized blocks.
Figure~\ref{fig:rq1_pareto} summarizes the trade-off between training speed and model quality; Dot size is proportional to
trainable parameters, with smaller dots indicating more
aggressive compression.
 
\paragraph{Finding 1.1: CDS improves efficiency with minimal quality degradation.}
Our CDS configuration ($L'\!=\!L/2$, $C\!=\!2$, intra-residual)
reduces trainable parameters by $26.6\%$ and improves training
throughput by $6.4\%$ while keeping offline AUC degradation within one basis point (0.01\%) of the production baseline. We further validate this configuration through a live A/B test on production traffic. CDS achieves a $+0.06\%$ online engagement gain with a $95\%$ confidence interval of $[-0.03\%, +0.15\%]$, indicating a neutral-to-positive online impact.

\paragraph{Finding 1.2: The intra-iteration residual is critical for CDS compression.}
Without the intra-iteration residual, CDS becomes significantly less stable and loses substantially more quality under the same parameter budget.
As shown in
Figure~\ref{fig:rq1_pareto}, the naive shared-block variant achieves similar throughput improvement but drops offline AUC by $13$ bps and produces a negative online engagement estimate ($-0.09\%$ with $95\%$ CI $[-0.18\%, +0.00\%]$). In contrast, CDS with intra-iteration residual reduces offline AUC by only $1$ bp while maintaining neutral-to-positive online performance. These results show that the residual connection is necessary for stable and effective CDS compression.

\paragraph{Finding 1.3: CDS remains effective under more aggressive compression.}
We further evaluate a more aggressively compressed CDS configuration
($L'\!=\!L/6$, $C\!=\!6$, intra-iteration residual). This setting reduces trainable
parameters by $44.4\%$ and improves training throughput by $12.3\%$,
at the cost of $9$ bps offline AUC drop and a slightly negative online
estimate ($-0.04\%$ with $95\%$ CI $[-0.15\%, +0.07\%]$).
Across all evaluated operating points, CDS with intra-iteration residual consistently achieves a better quality-efficiency trade-off than the naive shared-block variant.

\subsection{RQ2: The Efficiency of Chain Dimension Scaling}
\label{sec:experiments:rq2}
RQ1 demonstrated that Chain Dimension Scaling (CDS) improves parameter utilization, we investigate whether it can act as a primary scaling axis to improve model quality under strict resource constraints. 

\begin{figure}[t]
  \centering
  \includegraphics[width=\linewidth]{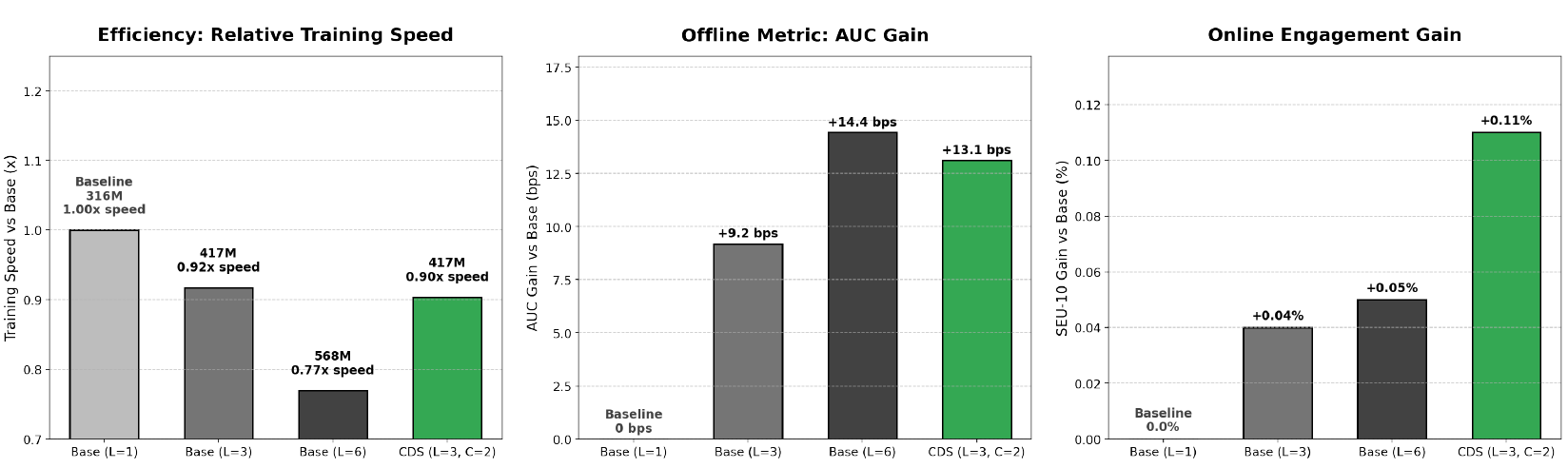}
  \caption[Efficiency vs Quality Scaling]{Comparison of standard Depth Scaling vs. Chain Dimension Scaling (CDS). The left panel shows the relative training speed and parameter footprint. The middle and right panels show the corresponding offline AUC gains and online engagement gains relative to the $L=1$ baseline. CDS ($L=3, C=2$) retains the efficiency of a 3-layer model while achieving quality metrics competitive with a 6-layer model.}  
  \label{fig:rq2_scaling}
\end{figure}

To evaluate this, we compare CDS against standard depth scaling. For standard scaling, we evaluate baselines with independent feature-interaction layers of depth $L \in \{1, 3, 6\}$ (with chain length $C=1$). For our approach, we apply CDS to the $L=3$ baseline by increasing the chain length to $C=2$. This CDS configuration ($L=3, C=2$) achieves an effective computation depth of $6$, allowing us to directly compare it against the standard $L=6$ model. Figure~\ref{fig:rq2_scaling} summarizes the trade-offs across training speed, offline AUC, and online engagement.

\paragraph{Finding 2.1: CDS scales quality without inflating the parameter footprint.}
As shown in the left panel of Figure~\ref{fig:rq2_scaling}, standard depth scaling from $L=1$ to $L=6$ drastically inflates the model size (from 316M to 568M parameters) and decreases training speed to $0.77\times$. Conversely, scaling the chain dimension from $C=1$ to $C=2$ on the $L=3$ model requires exactly zero additional parameters (remaining at 417M) and incurs only a marginal drop in training speed ($0.92\times \rightarrow 0.90\times$). Despite adding no new parameters, this chain scaling yields a substantial $+3.0$ bps improvement in AUC and a $+0.07\%$ absolute improvement in online engagement over the standard $L=3$ model.

\paragraph{Finding 2.2: CDS is a highly resource-efficient alternative to standard depth scaling.}
When comparing models with an equal effective depth of 6, the CDS model ($L=3, C=2$) performs remarkably close to the heavily parameterized baseline ($L=6, C=1$). While operating with roughly 150M fewer parameters and a significantly higher training speed, CDS achieves an offline AUC within 1 bps of the $L=6$ baseline ($+10.0$ bps vs $+11.0$ bps) and actually outperforms it slightly in online engagement ($+0.11\%$ vs $+0.05\%$). This confirms that the chain dimension provides a highly efficient path for scaling model capacity, successfully decoupling computation depth from parameter scale.

\subsection{RQ3: Which Residual Design Matters?}
\label{sec:experiments:rq3}
Section~\ref{sec:method:residual} introduces three residual designs for
CDS: no residual, intra-iteration residual, and combined intra+inter residual. In this section, we evaluate their impact on optimization stability and ranking quality.

\begin{figure}[t]
  \centering
  \includegraphics[width=0.65\linewidth]{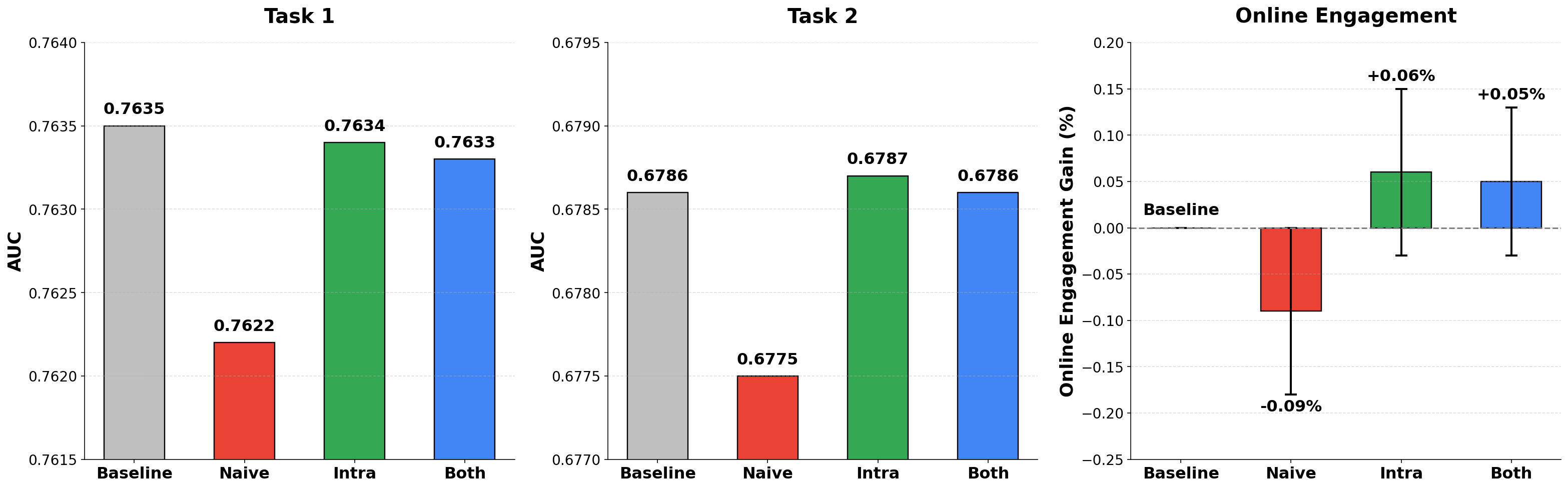}
  \caption{Residual design ablation within CDS. \textbf{Baseline}:
  production model with $L$ independently parameterized blocks (no
  CDS). \textbf{Naive}: CDS without residual (Eq.~\ref{eq:cds-forward}
  applied as written). \textbf{Intra}: CDS with the intra-iteration
  residual (Eq.~\ref{eq:intra-skip}). \textbf{Both}: CDS with intra
  and inter residuals combined. All CDS variants use $L'\!=\!L/2$ and
  $C\!=\!2$. Left and middle: absolute AUC on two representative
  classification tasks. Right: online engagement delta with $95\%$
  confidence intervals from live A/B tests.}
  \label{fig:rq3_residual}
  \vspace{-2mm}
\end{figure}

We fix the CDS compression configuration ($L' = L/2$, $C = 2$) and vary only the residual design. Figure~\ref{fig:rq3_residual} reports results on two representative offline ranking tasks together with online engagement metrics from live A/B tests.

\paragraph{Finding 3.1: CDS without residual connection suffers noticeable quality degradation.}
The Naive CDS variantwithout residual connections drops approximately $13$ bps
on the first offline task (AUC $0.7622$ vs.\ baseline $0.7635$) and
$11$ bps on the second ($0.6775$ vs.\ $0.6786$). The corresponding
online A/B test shows a $-0.09\%$ point estimate with the $95\%$ confidence interval of $[-0.18\%, +0.00\%]$.

These results suggest that parameter sharing alone is insufficient for stable optimization at longer chain lengths.
 
\paragraph{Finding 3.2: The intra-iteration residual largely recovers baseline quality.}
Adding the intra-iteration residual (Eq.~\ref{eq:intra-skip}) substantially improves optimization behavior. With the residual connection, CDS remains within $1$ bp of the baseline on the first task (AUC $0.7634$
vs.\ $0.7635$) and slightly improves the baseline on the
second task($0.6787$ vs.\ $0.6786$). The online A/B test shows a $+0.06\%$ engagement gain with neutral-to-positive confidence intervals. The residual connection provides a direct identity path across iterations, which improves gradient propagation and stabilizes training for shared iterative computation.
 
\paragraph{Finding 3.3: Additional inter-iteration residual connections provide limited benefit.}
We further evaluate a combined residual design that adds the inter-iteration residual on top of the intra-iteration residual. As shown in Figure~\ref{fig:rq3_residual}, the combined design produces nearly identical offline and online performance compared with the intra-iteration residual alone.

For example, the intra-iteration and intra-inter variant differ by at most $1$ bp on offline (Task 1 AUC
$0.7633$ vs.\ $0.7634$, Task 2 AUC $0.6786$ vs.\ $0.6787$, online
$+0.05\%$ vs.\ $+0.06\%$ with $95\%$ CIs heavily overlapping). Based on these results, all subsequent CDS experiments use the intra-iteration residual design.

\subsection{RQ4: Where the Efficiency Gain Comes From}
\label{sec:experiments:rq4}
\begin{figure}[t]
  \centering
  \begin{subfigure}[t]{0.8\linewidth}
    \centering
    \includegraphics[width=\linewidth]{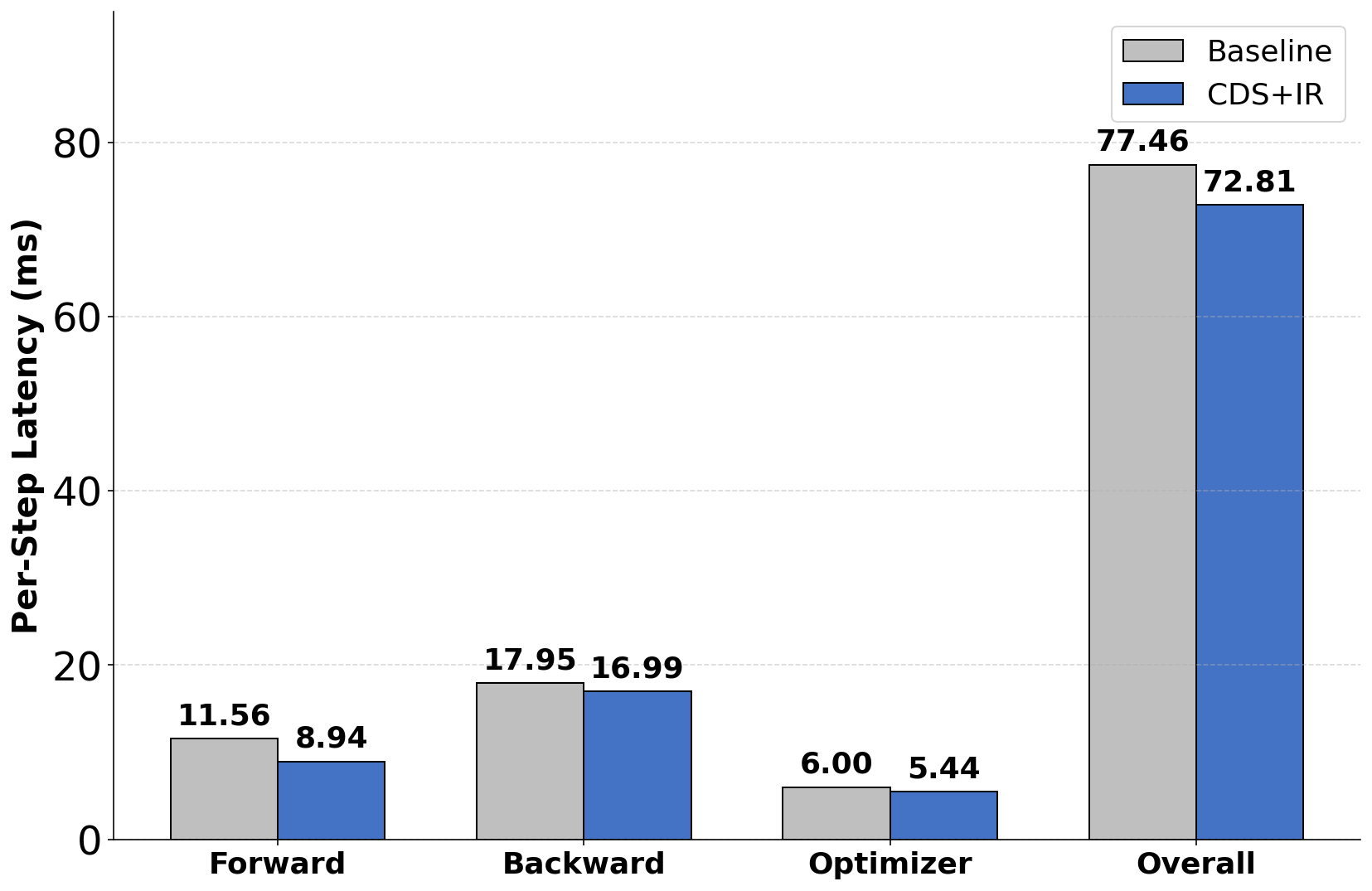}
    \caption{By training phase.}
    \label{fig:rq4_phase}
  \end{subfigure}
  \vspace{0.6em}
  \begin{subfigure}[t]{0.8\linewidth}
    \centering
    \includegraphics[width=\linewidth]{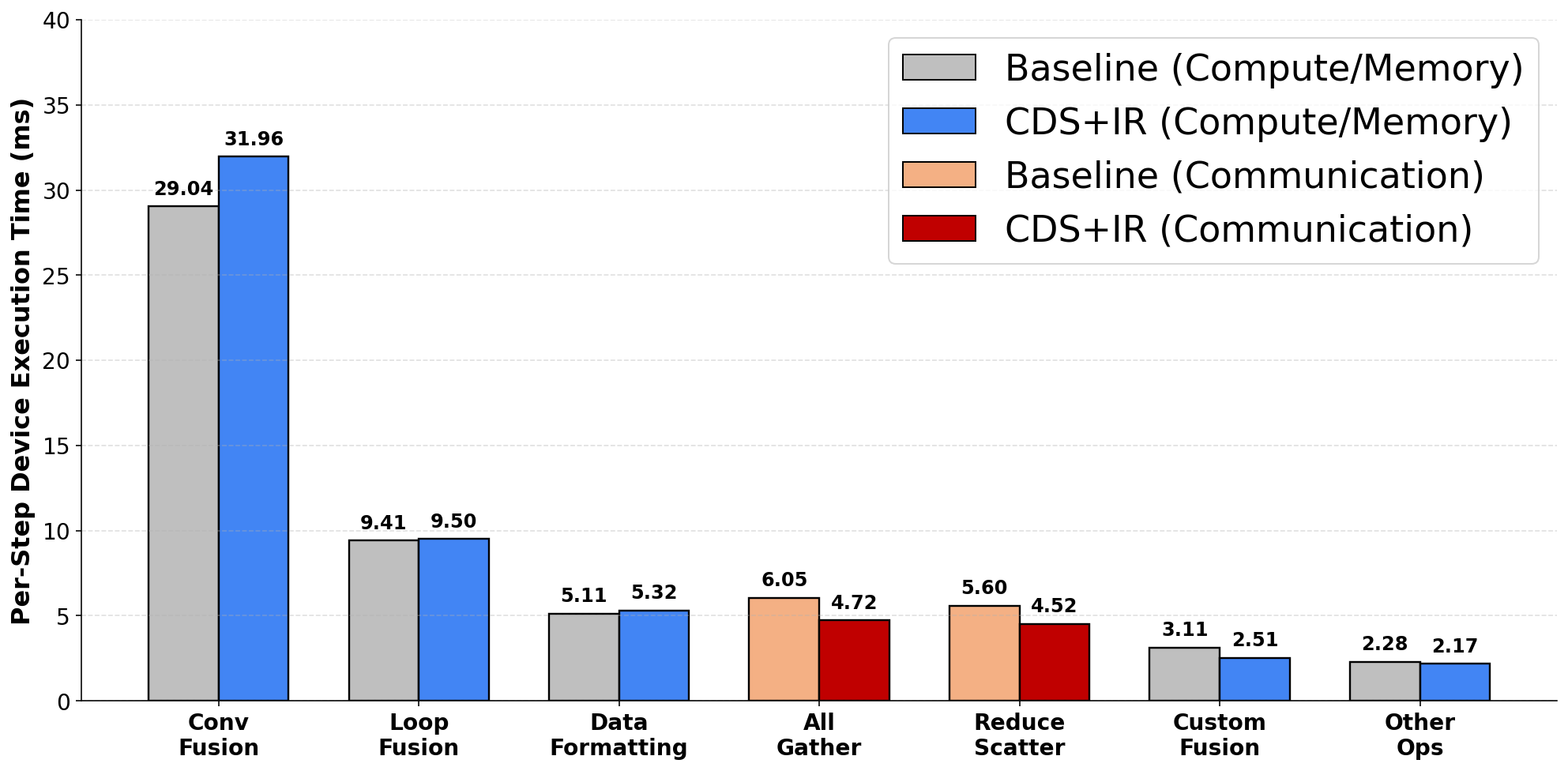}
    \caption{By device operation type.}
    \label{fig:rq4_ops}
  \end{subfigure}
  \caption{Wall-clock decomposition of the per-step training time
  for the production baseline vs.\ launched CDS, in milliseconds.
  \textbf{(a)} Phase-level breakdown: \emph{Forward} and
  \emph{Backward} are the interaction-block forward and backward
  passes; \emph{Optimizer} is the parameter update step;
  \emph{Overall} is the full training step (the remainder is
  occupied by ranker components unchanged by CDS: input feature
  layer, embedding lookups, shared projection, task towers).
  \textbf{(b)} Operation-level breakdown: compute and memory
  operations (grey/blue) vs.\ communication operations (orange/red:
  \emph{All Gather} for parameter unshard, \emph{Reduce Scatter}
  for gradient aggregation, both introduced by sharded
  data-parallel training).}
  \label{fig:rq4}
  \vspace{-2mm}
\end{figure}

The CDS compression improves training throughput by
$6.4\%$ while maintaining nearly unchanged ranking quality. In this section, we analyze where the efficiency gain comes from.
Figure~\ref{fig:rq4_phase} decomposes training time by training phase, while Figure~\ref{fig:rq4_ops} further breaks down execution time by device operation type.

\paragraph{Finding 4.1: Most of the wall-clock saving comes from the forward pass.}
Figure~\ref{fig:rq4_phase} shows that CDS reduces interaction-block forward time by $2.62$ ms, backward time by $0.96$ ms, and optimizer-step time by $0.56$ ms, resulting in an overall per-step reduction of approximately $4.14$ ms. Forward pass communication saving dominates the gain coming from. The remaining
step time is occupied by ranker components unchanged by CDS---input
pipeline, embedding lookups, shared projection, and task towers.

\paragraph{Finding 4.2: The efficiency gain mainly comes from reduced communication overhead, not
computation.}
Figure~\ref{fig:rq4_ops} shows that compute-related operations remain similar or slightly increase under CDS. For example, Conv Fusion time increases by $2.92$ ms due to additional iterative computation and residual operations.

In contrast, communication overhead decreases substantially. Parameter all-gather time decreases by $1.33$ ms, while gradient synchronization decreases by $1.08$ ms, resulting in a combined communication reduction of $2.41$ ms per step.

This behavior matches the mechanism described in Section~\ref{sec:method:cds}. CDS reduces the number of unique parameterized layers, which directly reduces per-layer communication overhead under sharded data-parallel training. Iterative reuse within the chain does not introduce additional parameter synchronization.

\paragraph{Finding 4.3: CDS also reduces optimizer overhead.}
The optimizer-step reduction of $0.56$ ms follows directly from the smaller parameter count. With fewer trainable parameters, each training step performs fewer optimizer updates and parameter-state operations.

Overall, these results show that CDS improves training efficiency primarily through reduced communication and optimizer overhead rather than reduced FLOPs. This mechanism is largely independent of the specific interaction block used in this work and should generalize to other parameter-shared architectures trained with sharded distributed training.

\section{Related Work}
\label{sec:related}
Two lines of prior work inform our setting: scaling efforts in
industrial ranking, and iterative computation with weight reuse in
both language modeling and recommendation. We position CDS against
each line.

\paragraph{Scaling Industrial Ranking Models} Industrial ranking has evolved from early DNN-based feature-interaction networks such as Wide \& Deep~\citep{cheng2016widedeep},
DeepFM~\citep{guo2017deepfm}, DCN~\citep{wang2021dcnv2}, and
AutoInt~\citep{song2019autoint}, through user-behavior models that
exploit sequential structure~\citep{zhou2018din,chen2019bst}, to
large-scale Transformer-based rankers that jointly model feature
interactions and sequential user behavior at industrial
scale~\citep{zhai2024hstu,han2025mtgr,zhang2025onetrans}.
In parallel, the community has sought to replicate the scaling laws of
large language models~\citep{kaplan2020scaling,hoffmann2022chinchilla}
inside the recommendation
domain~\citep{zhu2025rankmixer,khrylchenko2025billion,zhang2024wukong,jiang2026addressing} exploring three axes: depth scaling, width scaling, and data scaling. Multi-task learning, a feature of most production rankers, has been
advanced by gating-based architectures, including
MMoE~\citep{ma2018mmoe} and PLE~\citep{tang2020ple}. The empirical pattern is consistent across systems:
quality improves with scale, at the cost of parameter increase and
training-time growth. CDS is complementary to this body of work: it
introduces the chain dimension as a fourth, parameter-free axis inside
the same interaction module that recent industrial systems scale, and
can be composed with width, depth, or data scaling.

\paragraph{Iterative Computation and Weight Sharing}
Applying a single layer iteratively in place of stacking independent
layers has a long history in language modeling, from adaptive
computation time~\citep{graves2016act} and Universal
Transformers~\citep{dehghani2018universal} to ALBERT~\citep{lan2020albert}
and parameter-efficient transformers with shared
blocks~\citep{kitaev2020reformer}.
Recent work has revisited iterative
computation as a path to test-time scaling, latent reasoning, and
adaptive
depth~\citep{banino2021pondernet,saunshi2025looped,geiping2025recurrent,zhu2025loopedlm,prairie2026parcae,kohli2026looped,zhu2025vld};
related ideas have appeared inside Mixture-of-Experts via sequential
expert communication~\citep{wang2025coe}. Concurrently, this line has
begun to enter recommendation: the closest prior work,
LoopCTR~\citep{tang2026loopctr}, brings recursive weight sharing into
CTR prediction through a sandwich architecture with shared loop blocks
and process supervision, with strong results on public benchmarks.
 
CDS draws on the same conceptual foundation but differs on three
dimensions. \emph{Design:} we characterize CDS with intra-iteration residual inside a
multi-task production ranker serving live traffic, not benchmark or
controlled-study settings. \emph{Two-regime framing:} we frame the
chain dimension as a knob with two directions---compression and
quality-scaling---and the engineering payoff in our setting
comes from the compression direction, largely absent from prior
loop-style framings centered on inference efficiency or reasoning
capability. \emph{Communication-cost analysis:} our wall-clock decomposition
attributes the speedup to reduced per-layer all-gather and
reduce-scatter under sharded data-parallel
training~\citep{rajbhandari2020zero,zhao2023fsdp,xu2021gspmd}, not to
reduced FLOPs---an analysis, to our knowledge, not surfaced in prior work and one that should generalize to other
parameter-shared architectures.

\section{Conclusion}
\label{sec:conclusion}
We introduce Chain Dimension Scaling (CDS), a scaling strategy that increases effective computation depth through iterative parameter sharing. CDS applies feature-interaction blocks recurrently with shared parameters and a lightweight intra-chain residual, introducing chain length as a scaling axis orthogonal to width, depth, and data. CDS supports two operating regimes. In the compression regime of a production multi-task short-video ranker, CDS reduces trainable parameters by $26.6\%$ and improving training throughput by $6.4\%$
while remaining maintaining ranking quality. In the quality-scaling regime, increasing chain length consistently improves ranking quality under a fixed parameter budget.
Our analysis further shows that the efficiency gain mainly comes from reduced communication and optimizer overhead under sharded data-parallel training rather than reduced FLOPs. By reducing the number of unique parameterized layers, CDS lowers parameter synchronization cost during distributed training while preserving effective computation depth. More broadly, these results suggest that parameter-shared iterative computation can serve as a communication-efficient scaling strategy for large-scale recommendation systems. We expect the same perspective to apply to other parameter-shared architectures trained in distributed settings.

\bibliographystyle{plainnat}
\bibliography{references}

\end{document}